\documentclass[twocolumn]{article}
\usepackage{listings}
\usepackage{graphicx}
\usepackage{caption}
\usepackage{subfigure}
\usepackage{geometry}
\date{}
\usepackage{float}
\usepackage{amsmath}
\usepackage{mathtools}
\usepackage{indentfirst}
\usepackage{abstract}
\usepackage{verbatim }   
\usepackage{amssymb }  
\usepackage[utf8]{inputenc}
\usepackage{underscore}
\usepackage{indentfirst}
\usepackage{cite}
\usepackage{adjustbox}
\usepackage{blindtext}
\usepackage{makecell}
\usepackage{algorithm}
\usepackage{amsfonts,amssymb}

\usepackage{authblk}

\normalsize
\title{ ELA: Efficient Local Attention for Deep Convolutional Neural Networks }

\author[1,2,3]{Wei Xu}
\author[1]{Yi Wan}

\affil[1]{School of Information Science and Engineering, Lanzhou University}
\affil[2]{Qinghai Provincial Key Laboratory of IoT}
\affil[3]{Qinghai Normal University}

\begin{document}
\maketitle
\indent
\begin{abstract}
The attention mechanism has gained significant recognition in the field of computer vision due to its ability to effectively enhance the performance of deep neural networks. However, existing methods often struggle to effectively utilize spatial information or, if they do, they come at the cost of reducing channel dimensions or increasing the complexity of neural networks. In order to address these limitations, this paper introduces an Efficient Local Attention (ELA) method that achieves substantial performance improvements with a simple structure. By analyzing the limitations of the Coordinate Attention method, we identify the lack of generalization ability in Batch Normalization, the adverse effects of dimension reduction on channel attention, and the complexity of attention generation process. To overcome these challenges, we propose the incorporation of 1D convolution and Group Normalization feature enhancement techniques. This approach enables accurate localization of regions of interest by efficiently encoding two 1D positional feature maps without the need for dimension reduction, while allowing for a lightweight implementation. We carefully design three hyperparameters in ELA, resulting in four different versions: ELA-T, ELA-B, ELA-S, and ELA-L, to cater to the specific requirements of different visual tasks such as image classification, object detection and sementic segmentation. ELA can be seamlessly integrated into deep CNN networks such as ResNet, MobileNet, and DeepLab. Extensive evaluations on the ImageNet, MSCOCO, and Pascal VOC datasets demonstrate the superiority of the proposed ELA module over current state-of-the-art methods in all three aforementioned visual tasks.
\par

\end{abstract}

\begin{figure}[h]
	\centering
	\includegraphics[width=8cm,height=3.15cm]{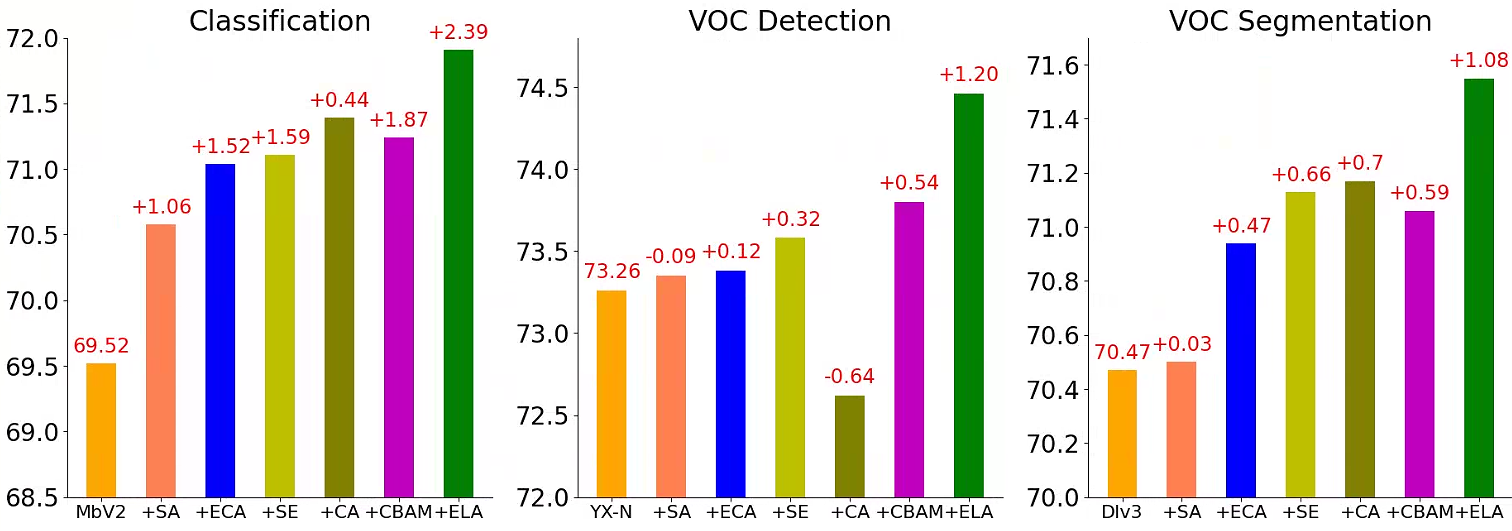}
	\caption{Performance comparison of multiple attention modules ( SA-Net \cite{zhang2021sa}, ECA-Net \cite{wang2020eca}, SE\cite{senet}, CA\cite{hou2021coordinate}, CBAM \cite{woo2018cbam}, and ELA) compared on three computer vision tasks.The y-axis labels from left to right are top-1 accuracy, AP and mean IoU, respectively. In the plot, “Mbv2” denotes MobileNetV2, “YX-N” denotes YOLOX-Nano, and “DLv3” represents DeepLabV3. Clearly, our approach demonstrates superior performance not only in ImageNet classification but also in VOC object detection and VOC semantic segmentation. }
	\label{fig1}
\end{figure}

\section*{1.Introduction}
Deep Convolutional Neural Networks (CNNs) have emerged as a vital research area in computer vision, yielding significant advancements in image classification, target detection, and semantic segmentation. Several noteworthy models, such as  AlexNet\cite{alexnet},  ResNet\cite{he2016resnet}, YOLO\cite{redmon2016you,li2022yolov6,wang2023yolov7}, and MobileNet\cite{howard2017mobilenets}, have contributed to this progress. While the Transformer has introduced numerous advancements \cite{2017attention,2018bert,vit}, deep convolutional neural networks possess their own advantageous inductive bias, enabling them to effectively learn from small and medium-sized datasets – a capability that the Transformer lacks. Consequently, the design of more efficient network architectures remains an important challenge for contemporary researchers to tackle \cite{2014vgg,ding2021repvgg,szegedy2015going,huang2017densely,li2017second,wang2023internimage}. In the realm of deep CNNs, the attention mechanism has been proposed as a means to simulate human cognitive behavior\cite{senet,xu2015neural, woo2018cbam,wang2020eca,hou2021coordinate,li2019selective,li2019spatial,hu2018gather}. This mechanism enables the neural network to focus on pertinent information while disregarding unimportant details, ultimately enhancing the network’s learning capabilities.
\par
One significant example is the SE block attention \cite{senet}, which leverages 2D global pooling to compress spatial dimensions into channel dimensions, facilitating enhanced feature learning. Nevertheless, SE block only considers encoding inter-channel information and neglects the spatial location information of the feature map. Though BAM\cite{park2018bam} and CBAM\cite{woo2018cbam} extract spatial attention, they fail to model the long-range dependencies\cite{romero2022towards} crucial for vision tasks, while also reducing the channel dimension of the input feature map. In response, the Coordinate Attention(CA)\cite{hou2021coordinate} method was developed, embedding spatial location information into channel attention and allowing mobile networks to capture long-range spatial interrelationships accurately. This improvement benefits various CNN architectures.However, CA also exhibits evident limitations stemming from its insufficient generalization ability and the adverse effects on channel dimensionality reduction.
\par
As we all know, the spatial dimension of an image contains crucial positional information. On the other hand, existing attention mechanisms either fail to effectively utilize this spatial information or do so at the cost of reducing the channel dimensionality. The focus of this paper is to address the following question: Can we learn spatial attention in a more efficient manner? This approach should allow us to obtain accurate location predictions in the spatial dimension without compromising the channel dimensionality of the input feature map, while also maintaining the lightweight nature of the attention module.
\par
To address this question, let’s revisit the CA mechanism\cite{hou2021coordinate}. The CA module is devised in two steps. In the first step, strip pooling\cite{hou2020strip}  is utilized to generate feature maps containing horizontal and vertical coordinate information for the spatial dimensions of the input tensor. In the second step, the aforementioned feature maps in both directions undergo two 2D convolutions, followed by Batch Normalization (BN) \cite{wu2018group}. and a non-linear activation function, yielding the final attention. It is evident that the design process of CA is relatively intricate, involving multiple separations and mergers of the feature maps in both directions. Although the two 2D convolutions enhance the coordinate information, they also reduce the channel dimension, causing a negative impact on the generated attention. Additionally, incorporating BN into CA introduces notable weaknesses. For example, excessively small mini-batch sizes can detrimentally affect the entire model and hinder its generalization ability. The results of the ablation experiments in Table\ref{tab2} and Table\ref{tab3} further support these observations.
\par
\begin{figure*}[t!]
	\centering
	\includegraphics[width=\linewidth]{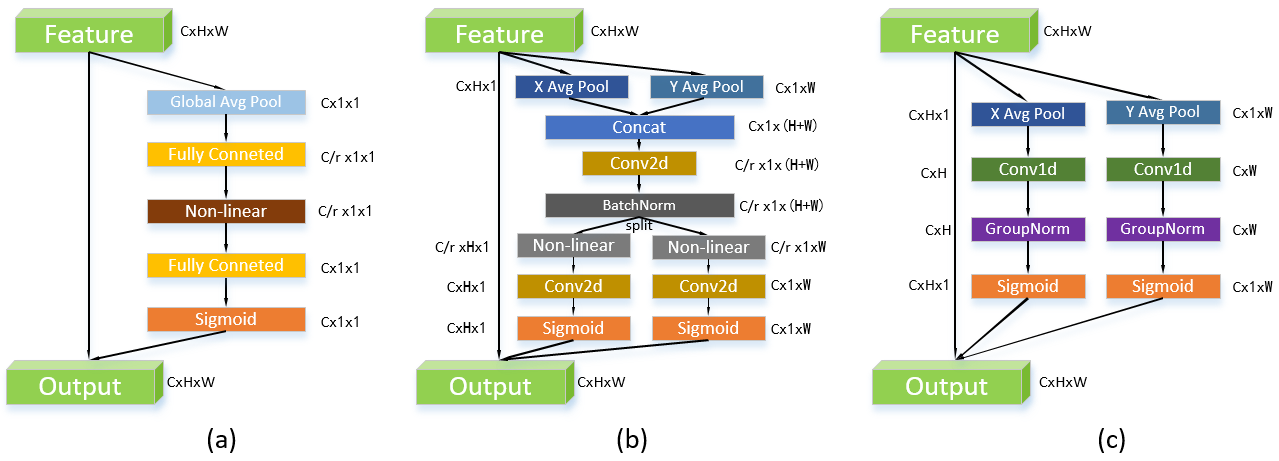}
	\caption{The schematic diagrams of Efficient Local Attention (ELA) (c) are compared with SE block \cite{senet} (a) and Coordinate Attention (CA) \cite{hou2021coordinate} (b). “X Avg Pool” and “Y Avg Pool” represent one-dimensional horizontal global pooling and one-dimensional vertical global pooling, respectively. From a structural perspective, the ELA appears to be significantly more lightweight compared to the CA, while also avoiding dimension reduction in the channel dimension.}
	\label{fig2}
\end{figure*}

Hence, this paper proposes the Efficient Local Attention (ELA) module for deep CNNs, which accurately captures the location of regions of interest, maintains the input feature map channels’ dimensionality, and preserves its lightweight characteristics, as demonstrated in Fig \ref{fig2}(c). Similar to CA, ELA employs strip pooling \cite{hou2020strip} in the spatial dimension to obtain feature vectors in both horizontal and vertical directions, maintaining a narrow kernel shape to capture long-range dependencies and prevent irrelevant regions from affecting label prediction, resulting in rich target location features in their respective directions. ELA processes the above feature vectors independently for each direction to obtain attention predictions, which are then combined using a product operation, ensuring accurate location information of the region of interest. Specifically, in the second step, 1D convolution is applied to locally interact with the two feature vectors separately, with the option to adjust the kernel size to indicate the coverage of the local interaction. The resulting feature vectors undergo grouping normalization (GN) \cite{wu2018group} and nonlinear activation functions to produce positional attention predictions for both directions. The final positional attention is obtained by multiplying the positional attention in both directions. Compared to 2D convolution, 1D convolution is better suited for processing sequential signals and is lighter and faster. GN demonstrates comparable performance and greater generalizability compared to BN.
\par
Table \ref{tab1} presents the key CNN attention modules, indicating their characteristics in terms of channel dimensionality reduction (DR), long-range dependency, and lightweight design (where lightweight models have fewer parameters than SE). It is evident from the table that our ELA excels across all three aspects. We evaluate the effectiveness of our approach through experimental results on datasets such as ImageNet \cite{deng2009imagenet}, Pascal VOC \cite{2015voc}, and MS COCO \cite{2014coco} (see table \ref{tab5}). The experimental findings demonstrate a 2.39\% improvement in the classification accuracy of our proposed method on ImageNet top-1. Furthermore, our method exhibits the most significant performance enhancement in target detection and semantic segmentation. Hence, our proposed ELA method proves to be more competitive than the currently most popular attention mechanism.
\par

\begin{table}[htbp]\small
    \centering
    \setlength{\tabcolsep}{2.5pt}
    \renewcommand{\arraystretch}{0.85}
    \caption{Comparison of existing attention method in terms of whether no channel dimensionality reduction (No DR) , long-range dependencies and less parameters than SE block \cite{senet} (indicated by lightweight) or not.}
    \begin{tabular}{ c | c | c | c }  
     \hline  
         Method       & No-DR         & \makecell{Long-range \\ dependencies}   & Lightweight \\
         \hline
          SE block   & $\times$     &  $\times$                              & -      \\

          CBAM       & $\times$     & $\times$                              & $\times$     \\

          ECA        & \checkmark   &  $\times$                              & \checkmark   \\

          CA         & $\times$     & \checkmark                             & $\times$      \\

          SA         & $\times$     & $\times$                                & \checkmark    \\

          ELA(Ours)  & \checkmark   & \checkmark                              & \checkmark      \\
          \hline
    \end{tabular}
    \label{tab1}
\end{table}

The contributions of this paper are summarized as follows.
(1) We analyze Coordinate Attention(CA)\cite{hou2021coordinate} and experimentally verify the detrimental effects of its BN and channel dimension reduction on the CNN architecture.  (2) Building upon the aforementioned analysis, we propose a lightweight and Efficient local attention (ELA) module. This module assists deep CNNs in accurately localizing the object of interest, significantly improving the overall performance of CNNs with only a marginal increase in parameters. (3) Extensive experimental results on popular datasets, including ImageNet, MS COCO, and Pascal VOC, demonstrate that our proposed method surpasses the current state-of-the-art attention methods in terms of performance while maintaining competitive model complexity.
\par

\section*{2. Related Work}
It is widely recognized that the attention mechanism plays a crucial role in enhancing deep CNNs. The SE block\cite{senet} was the first successful attempt at incorporating an attention for learning channel attention. Subsequently, attention mechanisms have made significant advancements in these two directions: (1) aggregating channel features only; (2) integrating channel features with spatial features.
\par
Specifically, CBAM \cite{woo2018cbam} simultaneously utilizes average pooling and max pooling to combine features along both the channel dimension and spatial dimension. Meanwhile, GCNet\cite{cao2019gcnet} is a lightweight attention network that leverages techniques such as self-attention mechanisms \cite{2017attention}, nonlocal networks \cite{wang2018non}, and squeeze-excitation networks \cite{senet}. SA-Net\cite{zhang2021sa} builds upon the combination of Spatial Attention and Channel Attention and introduces feature grouping and channel substitution to achieve a lightweight attention mechanism. CBAM, GCNet, and SA-Net all combine Spatial Attention and Channel Attention. GSoP\cite{gao2019gsop} introduces second-order pooling to enable higher-order statistical modeling of the entire image, thereby enhancing the nonlinear modeling capability of deep convolutional networks. On the other hand, ECA-Net\cite{wang2020eca} utilizes 1D convolution to generate channel attention weights, significantly reducing the modeling complexity compared to SE block. GSoP and ECA-Net both fall under the category of channel-enhanced aggregation approaches.
\par
However, in the attention networks mentioned above, there is either a lack of long-range dependency in the spatial dimension or a downsizing of the channel dimension. The absence of long-range spatial dependence makes it challenging to accurately localize spatial objects of interest and obtain location information about important objects. While reducing the complexity of the model is possible through channel dimensionality reduction, it also disrupts the direct correspondence between the channels and their weights. To address these limitations, our proposed ELA method effectively captures long-range spatial dependency and eliminates the negative impact caused by channel downsizing on attention networks.
\par

\section*{3. Method}
The Efficient Local Attention module serves as a computational unit aimed at enhancing the accurate identification of regions of interest or significant object locations in deep CNNs. To provide a clear explanation of our proposed efficient localization attention (ELA) method, this section begins by restating the two steps involved in CA \cite{hou2021coordinate}. Subsequently, we examine the impact of BN and channel downscaling and empirically evaluate the application of CA to some small models. These findings lay the foundation for introducing ELA. We then delve into the process of constructing ELA and design four versions by combining hyperparameters of 1D Convolution and GN within ELA. Finally, we employ the Grad-cam \cite{2017grad} method for visualization, further illustrating the effectiveness of ELA. Figure \ref{fig2}(b) provides an overview of the overall structure of ELA.
\par
\subsection*{3.1 Revisit Coordinate Attention}
\subsubsection*{3.1.1 Coordinate Attention}
The CA consists of two main steps: Coordinate Information Embedding and Coordinate Attention Generation. In the first step, the authors propose a smart method of capturing long-range spatial dependencies by utilizing strip pooling instead of spatial global pooling, which is a well-thought design decision.
\par
Let’s consider the output of a convolutional block, denoted as $\mathbb{R}^{H\times W\times C}$, representing the height, width, and channel dimensions (i.e., the number of convolutional kernels) respectively. To apply strip pooling \cite{hou2020strip}, we perform average pooling on each channel within two spatial ranges: (H,1) along the horizontal direction and (1,W) along the vertical direction. This results in an output representation for the c-th channel at height h, as well as an output representation for the c-th channel at width w. These can be mathematically expressed using Eq\ref{eq1} and Eq\ref{eq2}.
\par
\begin{equation}
  z^h_c(h) = \frac{1}{ H}{\sum_{0\leq i<H} x_c{(h,i)}},
    \label{eq1}
\end{equation}

\begin{equation}
  z^w_c(w) = \frac{1}{ W}{\sum_{0\leq j<W} x_c{(j,w)}},
    \label{eq2}
\end{equation}
In the second step, the feature maps generated by Eq \ref{eq1} and Eq \ref{eq2} are aggregated to become a new feature map. It is then sent to the shared transformation function $F_1$ (which is a 2D convolution) and BN. They can be represented as follows.
\par
\begin{equation}
  f = \delta(BN(F_1([z^h,z^w]))),
    \label{eq3}
\end{equation}

In the above description, the cascade operation along the spatial dimension is denoted by $[-;-]$. $\delta$ represents a non-linear activation function. The intermediate feature map, denoted as $\mathbb{R}^{C/r{\times(H+W)}}$ , is obtained after encoding horizontally and vertically. Subsequently, $f^h \in {\mathbb{R}^{C/r \times H}}$, $f^w \in {\mathbb{R}^{C/r \times W}}$, along the spatial dimension. Additionally, the other two $ 1\times 1 $ convolutional transforms, denoted as$F_h$, $F_w$, are applied to generate tensors with the same number of channels as the input $X$.
\par

\begin{equation}
  g_c^h = \sigma(F_h(f^h)),
    \label{eq4}
\end{equation}
\begin{equation}
  g_c^w = \sigma(F_w(f^w)),
    \label{eq5}
\end{equation}
The sigmoid function $\sigma$  is utilized in this context as well. In order to decrease the complexity of the overhead module, the number of channels in $f$ is typically reduced by an appropriate reduction rate, such as $32$. The resulting outputs, $g_c^h$ and $g_c^w$, are expanded and employed as attention weights, corresponding to the horizontal and vertical directions respectively. Ultimately, the outputs of the CA module can be represented as $Y$.
\par
\begin{equation}
  y_c(i,j) = x_c(i,j)\times{g_c^h(i)}\times{g_c^w(j)},
    \label{eq6}
\end{equation}
Upon observing Eqs \ref{eq3}, \ref{eq4}, and \ref{eq5}, we can note that the channel dimensionality reduction aims to decrease the complexity of the model. However, it results in an indirect relationship between the channels and their corresponding weights, which can lead to adverse effects on the overall attention prediction. Additionally, it is important to highlight that BN does not facilitate effective generalization ability for CA. In the subsequent sections, we will delve into further details regarding these observations.
\par
\subsubsection*{3.1.2 Shortcomings of Coordinate Attention}
As stated in the research conducted by Wu et al. (2018) \cite{wu2018group}, Batch Normalization (BN) heavily relies on the mini-batch size. When the mini-batch is too small, the mean and variance calculated by BN may not adequately represent the entire dataset, potentially impairing the overall model performance. The Coordinate information embedding obtained from Eq. \ref{eq1} and Eq. \ref{eq2} represents sequential information within each channel dimension. Placing BN in a network that handles serialized data is not optimal, particularly for the CA approach. Consequently, CA could negatively impact smaller network architectures. Conversely, when Group Normalization (GN) is used as a replacement for BN within CA and incorporated into smaller network architectures, notable improvements in performance immediately occur.
Furthermore, a thorough analysis of the structure of CA can expose additional challenges. At the onset of its second step, the feature maps $z_h$ and $z_w$ are concatenated into a new feature map and subsequently encoded. However, the feature maps  $z_h$ and $z_w$ in both directions possess distinctive characteristics. Consequently, once merged and their features captured, the mutual influence at their respective connections may undermine the accuracy of attention prediction in each direction.
\par
Table \ref{tab2} and table \ref{tab3} clearly demonstrate the performance degradation of CA with BN. For instance, the Top 1 accuracy of ResNet18 decreases by $1.12\%-0.09\%$, and the performance of YOLOX-Nano drops by $1.57\%-0.64\%$. Whereas, upon replacing BN with GN, the Top 1 accuracy of ResNet18 improves by $0.32\%-0.44\%$, and the performance of YOLOX-Nano is raised by $0.51\%-0.70\%$. Hence, when working with small models, the utilization of BN undermines the generalization performance of CA. Moreover, it is worth noting that CA with BN introduces significant uncertainty to the performance due to changes in the dimensionality reduction factor, lacking the regularity demonstrated by larger models.
\par

\begin{table}[h]\small
    \centering
    \setlength{\tabcolsep}{2.5pt}
    \renewcommand{\arraystretch}{0.85}
    \caption{We conducted a comparative experiment on ImageNet, utilizing ResNet-18 as the backbone model, to evaluate Coordinate Attention and its modification scheme. In this experiment, the dimensionality reduction factor, denoted as “r”, was taken into account. The results demonstrate a notable improvement in the model’s performance when GN was employed as a substitute for BN. Additionally, it is evident that the downscaling factor has a significant impact on the model’s performance.}
    \begin{tabular}{ c | c | c | c  }  
     \hline  
         Model      & top-1                    & Params(MB)            & GFLOPs(G)        \\
         \hline
          ResNet18  &  69.86                  & 11.690            & 1.824             \\

          +CA(r=32)  &\textbf{69.77}            & \textbf{11.788}  & \textbf{1.825}    \\

          +CA(r=24)  &  68.74                 &  11.816             & 1.826             \\

          +CA(r=16)  &  69.00                 &  11.886              & 1.827            \\
          \hline
          +CA_gn(r=32)  &  70.18              & \textbf{11.795}       & \textbf{1.826}   \\
          +CA_gn(r=24)  &  70.21            & 11.826            & \textbf{1.826}        \\
          +CA_gn(r=16)  &  \textbf{70.30}        & 11.888              & 1.827           \\
          \hline
          ResNet50  &  75.83                & 25.557               & 4.134              \\

          +CA(r=32)  &  76.35                & \textbf{26.080}       & \textbf{4.141}    \\
          +CA(r=24)  &  76.42             & 26.249             & 4.143                \\
          +CA(r=16)  &  \textbf{76.64}         & 26.603             & 4.148              \\

          \hline
    \end{tabular}
    \label{tab2}
\end{table}

\begin{table}[h] 
    \centering
    \caption{Comparative experiments were conducted on the YOLOX-Nano architecture, specifically examining Coordinate Attention and its modified scheme. The evaluation was performed on the Pascal VOC2007 dataset. In these experiments, the dimensionality reduction factor, denoted as “r”, was varied. The results clearly indicate that substituting BN with GN significantly enhances the model’s performance. Moreover, it is evident that the downscaling factor has a noticeable effect on the model’s performance.}
    \begin{tabular}{ c | c | c | c }  
     \hline  
         Method          & mAP             & \#P(MB)       & GFLOPs(G) \\
         \hline
          YOLOX-nano     &  73.26          & $3.43$            & 1.285      \\
          +CA(r=16)       &  \textbf{72.62}  &$3.61$              & 1.287    \\

          +CA(r=24)       &  71.57           & $3.55$              &1.286   \\

          +CA(r=32)      &  72.53          & \textbf{3.52}       &\textbf{1.286}     \\
          \hline
          +CA_gn(r=16)   &  73.77       & 3.62                 & 1.287       \\
          +CA_gn(r=24)   &  73.30          & 3.56                & 1.286          \\
          +CA_gn(r=32)   &  73.96         & \textbf{3.52}       &\textbf{1.286}     \\
          \hline

        \end{tabular}
    \label{tab3}
\end{table}

Furthermore, in Figure \ref{fig2}(b), it was observed that in the Coordinate Attention Generation process, 2D convolutions were employed twice. These convolutions enhanced the Coordinate Information but resulted in dimension reduction of the channels. While this process reduces the model’s complexity, it introduces noticeable side effects to the generation of attention.
\par
Even though MobileNetV2 \cite{2018mobilenetv2} has only 3.5MB parameters, why does CA lead to significant performance improvements in MobileNetV2? While MobileNetV2 has fewer parameters compared to ResNet18, which consists of only 18 convolutional layers and fully connected layers, and is considered a smaller model, MobileNetV2-1.0, on the other hand, has as many as 57 convolutional layers, surpassing even ResNet50. Additionally, MobileNetV2 is typically trained with a $batch\_size$ of 256, allowing the application of CA to avoid the detrimental effects of BN and effectively leverage its benefits. As we know, according to \cite{wu2018group}, the significant reduction in computational complexity and parameters in MobileNetV2-1.0 is achieved through the use of Depthwise Separable Convolution \cite{chollet2017xception}. If normal convolution is utilized, the number of parameters will significantly increase.
\par
\subsection*{3.2 Efficient Local Attention}
The CA method shows significant improvements in accuracy, especially for deeper networks, by utilizing strip pooling to capture long-range dependencies in spatial dimensions. Building upon our earlier analysis, it becomes evident that BN hampers the generalization ability of CA, while GN addresses these deficiencies. The localization information embeddings derived from Eq. \ref{eq1} and Eq. \ref{eq2} are sequential signals within the channel. Consequently, it is typically more appropriate to employ 1D convolution rather than 2D convolution for processing these sequential signals. Not only is 1D convolution adept at handling sequence signals, but it is also more lightweight compared to 2D convolution. In the case of CA, although 2D convolution is employed twice, it utilizes $1 \times 1$ convolution kernels, which limits the extraction capabilities of the features. Therefore, we employ 1D convolution with kernel sizes of 5 or 7, which effectively enhances the interaction capabilities of the localization information embedding. This modification enables the entire ELA  to accurately locate regions of interest.
\par
Based on the localization information embedding obtained from Eq. \ref{eq1} and Eq. \ref{eq2}, our ELA  employs a novel encoding method to generate a precise location attention map. The detailed description of this process is provided below.
\par
The $z_h$ and $z_w$, obtained from Eq. \ref{eq1} and Eq. \ref{eq2} capture not only global sensory fields but also precise location information. To leverage these features effectively, we have devised straightforward processing methods. We apply 1D convolution to enhance the positional information in both horizontal and vertical directions. Subsequently, we utilize GN (denotes as $G_n$)  to process the enhanced positional information, resulting in the representation of positional attention in the horizontal and vertical directions, as described in Eq. \ref{eq7} and Eq. \ref{eq8}.
\par
\begin{equation}
  y^h = \sigma(G_n(F_h(z_h))),
    \label{eq7}
\end{equation}
\begin{equation}
  y^w = \sigma(G_n(F_w(z_w))),
    \label{eq8}
\end{equation}

In the above description, we represent the nonlinear activation function as $\sigma$ and denote the 1D convolutions as  $F_h$ and $F_w$. We choose to set the convolution kernels of $F_h$ and $F_w$ to either 5 or 7. Generally, a convolution kernel of 7 tends to perform better, albeit with a slightly larger number of parameters. To strike a balance between performance and parameter count, the $groups$ of the 1D convolution are typically chosen to be either $in\_planes$ or $in\_planes/8$. The resulting representations of location attention in the horizontal and vertical directions are denoted as $y^h$ and $y^w$, respectively. Finally, we can obtain the output of the ELA module, denoted as $Y$, by applying Eq. \ref{eq9}.
\par
\begin{equation}
  Y = x_c \times y^h \times y^w,
    \label{eq9}
\end{equation}

\subsection*{3.3 Multiple ELA version settings}

According to Eq \ref{eq3} and Eq \ref{eq4}, our ELA involves three parameters: $kernel\_size$ and $groups$ for 1D convolution, and $num\_group$ for GN. To enhance the performance of the CNNs, it is crucial to set these parameters effectively. We aim to strike a balance between the performance and complexity of the ELA . In our experiments, we use ResNet-18 and ResNet-50 as backbones and incorporate the ELA to determine reasonable parameter values. For the $num\_group$ parameter of GN, we refer to \cite{wu2018group} and choose values of 16 and 32, respectively. 1D Convolution enables the capture of location information within the interactions. Generally, a larger $kernel\_size$ in 1D Convolution provides broader coverage of location information, leading to slightly better performance results. However, it also increases the complexity of the ELA . In our experiments, we evaluate $kernel\_size$ values of 5 and 7, and find that $kernel\_size=7$ offers improved performance, particularly for ResNet-50. Regarding the $groups$ parameter of 1D convolution, we explore two schemes: Depthwise Convolution (with $groups=in\_channels$) and Group Convolution (with $groups= in\_channels/8$). According to the results presented in Table \ref{tab3}, in most cases, the utilization of Group Convolution ($groups= in\_channels/8$) outperforms Depthwise Convolution ($groups= in\_channels$). Additionally, we observe that for ResNet-18, apart from  $kernel\_size=7$, a $num\_group$ value of 16 yields better results.

\begin{table}[htbp]\small
    \centering
    \setlength{\tabcolsep}{2.5pt}
    \renewcommand{\arraystretch}{1}
    \caption{Test the performance of ELA with different hyperparameters on ResNet-18 and ResNet-50. Evaluate the models using the ImageNet dataset, with k7 and k5 denoting $kernel\_size$ values of 7 and 5, respectively. g and g8 represent $groups=in\_channels$ and $groups=in\_channels/8$, respectively. ng16 and ng32 correspond to $num\_group$ values of 16 and 32, respectively.}
    \begin{tabular}{ c | c | c | c }  
     \hline  
         Method                     & Top-1                  & P(MB)               & GFLOPs(G) \\
         \hline
          ResNet18                  &  69.86                & 11.690             & 1.824  \\
          +ELA-k7-g8-ng16(L)         &  70.60                & 11.851             & 1.830     \\
          +ELA-k7-g-ng16(B)          &  70.44                & 11.710	        & 1.825   \\
          +ELA-k7-g8-ng32             & 70.77	                & 11.851             & 1.830    \\
          +ELA-k7-g-ng32	            & 70.52	                & 11.710            & 1.825   \\
          +ELA-k5-g8-ng16(S)       	& \textbf{70.79}	    & 11.805	        & 1.828   \\
          +ELA-k5-g–ng16	            & 70.51		            & \textbf{11.704}   & \textbf{1.825}   \\
          +ELA-k5-g8-ng32(T)       	& 70.68	 	            & 11.805             & 1.828   \\
          +ELA-k5-g–ng32	            & 70.28	                & \textbf{11.704}	 & 1.825   \\
          \hline
          ResNet50	               & 75.83		           & 25.557	              & 4.134   \\
          +ELA-k5-g8-ng32	       & 76.37		           & 25.713	              & 4.139   \\
          +ELA-k5-g8-ng16(S)        & 76.59	        	  & 25.713	             & 4.139   \\
          +ELA-k7-g8-ng32	       & 76.54		           & 25.776              & 4.141   \\
          +ELA-k7-g-ng16(B)         & \textbf{76.63}	      & \textbf{25.565}	     & \textbf{4.135}   \\
          +ELA-k7-g8-ng16(L)        & 76.56	               & 25.776	              & 4.141   \\
          \hline
    \end{tabular}
    \label{tab4}
\end{table}

\par
To optimize the performance of ELA while also considering the parameter count, we introduce four schemes: ELA-Tiny(ELA-T), ELA-Base(ELA-B), ELA-Small(ELA-S), and ELA-Large(ELA-L). We define the parameter configuration of ELA-T as $kernel\_size=5$, $groups=in\_channels$, $num\_group=32$; define the parameter configuration of ELA-B as $kernel\_size=7$, $groups=in\_channels$, $num\_group=16$; and the parameter configuration of ELA-S is $kernel\_size=5$, $groups=in\_channels/8$, $num\_group=16$. The parameter configuration of ELA-L is $kernel\_size=7$, $groups=in\_channels/8$, $num\_group=16$. ELA-T and ELA-B are designed to be lightweight, making them ideal for CNN architectures with fewer network layers or lightweight networks. On the other hand, ELA-B and ELA-S perform best on networks with deeper structures. Additionally, ELA-L is particularly suitable for large networks. It is worth noting that even ELA-L, despite having fewer parameters than the lightest of the CA methods (r=32), still delivers impressive results.
\par

\subsection*{3.4 Visualization}
\begin{figure}[h]
	\centering
	\includegraphics[width=8cm,height=5.5cm]{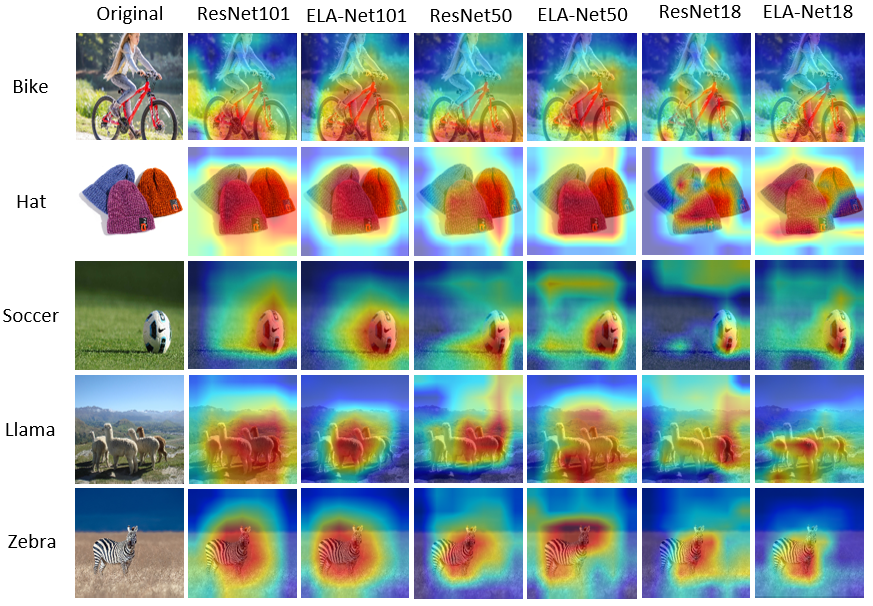}
	\caption{The visualization examples generated by GradCAM \cite{2017grad} depict the use of “layer4.2” across all target layers. The results clearly demonstrate that our localization attention module (ELA) effectively localizes the objects of interest with a high level of accuracy.}
	\label{fig3}
\end{figure}

In order to assess the effectiveness of the ELA method, we conducted two sets of experiments on ImageNet: ResNet \cite{he2016resnet} (without the attention module) and ELA-ResNet (with ELA). To evaluate the performance, we utilized five images for testing. By generating visual heat maps using GradCAM \cite{2017grad}, we presented the results for both sets of models in layer4.2 (the final bottleneck of the last stage). Figure \ref{fig3} illustrates that our proposed ELA module successfully guides the entire network to focus more precisely on the relevant regions of object details. This demonstration highlights the efficacy of the ELA module in enhancing classification accuracy.
\par

\subsection*{3.5 Implementation}
\begin{figure}[htbp]
	\centering
	\includegraphics[width=8cm,height=5cm]{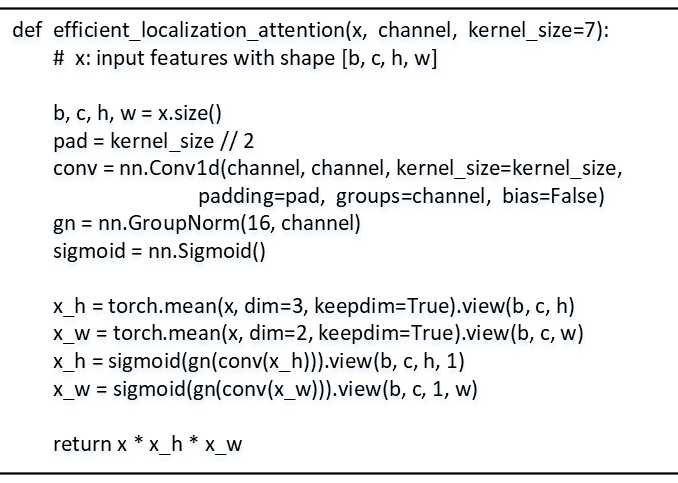}
	\caption{PyTorch code for our proposed ELA-B module}
	\label{fig4}
\end{figure}

Figure \ref{fig2}(b) presents an overview of our ELA . This module can be seamlessly integrated into deep CNNs that have the same configuration as CA \cite{hou2021coordinate}. Implementing the ELA in PyTorch or TensorFlow involves writing only a small amount of code, which supports automatic differentiation. To demonstrate this, we provide the PyTorch code for our ELA-B in Figure \ref{fig4}.
\par

\section*{4 Experiment}
In this section, we evaluate the performance of our proposed method on large-scale image classification, object detection, and semantic segmentation tasks using the ImageNet \cite{deng2009imagenet}, MS COCO \cite{2014coco}, and VOC2007/2012 \cite{2015voc} datasets, respectively. Specifically, we provide an overview of our experimental setup, compare our proposed ELA method with state-of-the-art counterpart modules, and present comparative results of the ELA method and other counterparts in terms of object detection and semantic segmentation.
\par
\subsection*{4.1 Experiment details}
All experiments were performed on the PyTorch toolbox \cite{2019PyTorch} and training was done using NVIDIA GPUs. To evaluate the classification effectiveness of our ELA module on ImageNet, we used four of the CNNs as the backbone, including  MobileNetV2 \cite{2018mobilenetv2},ResNet-18 \cite{he2016resnet}, ResNet-50 \cite{he2016resnet} and ResNet-101 \cite{he2016resnet} .
\par
For MobileNetV2, the mini-batch size is set to 256, the SGD optimizer is used, the momentum is all 0.9, and the weight decay is set to $4\times10^{-5}$. We learn the rate decay strategy by cosine annealing with an initial learning rate of $0.025$, and all the models are trained for a total of 200 epochs. For the data augmentation we use the same method as for MobileNetV2 \cite{2018mobilenetv2}.  We give classification results on the ImageNet dataset .
\par
For all three ResNet networks, we adopted the same data augmentation and hyperparameter settings as described in \cite{he2016resnet}. Specifically, the input images were randomly cropped to $224
\times224$ and horizontally flipped randomly. The network parameters were optimized using stochastic gradient descent (SGD) with a weight decay of $1\times e^{-4}$ and a momentum of $0.99$.  All models were trained for a total of 90 epochs, with the initial learning rate set to either 0.1 and 0.05. The learning rate was reduced by a factor of 10 every 30 epochs.
\par

\begin{table*}[h] 
    \centering
    \caption{When compared to the state-of-the-art attention module on ImageNet, it was consistently observed that if the modules have reduced channel dimensions, its scaling factor is set to 32 by default.}
    \begin{tabular}{ c | c | c | c | c| c }  
     \hline  
         Attention Methods  &	Bankbones             & Top-1(\%)       & Top-5(\%)           & Params(M)           &	GFLOPs \\

          \hline
          MobileNetV2-1.0	& MobileNetV2-1.0	   & 69.52	             & 89.21 	          & 3.50	            & 327.56M       \\
          +CA(r=32)		    & MobileNetV2-1.0      & 71.39          	 & 90.57	          & 3.95	            & 337.46M       \\
          +CA(r=24)		    & MobileNetV2-1.0	   & 71.56          	 & 90.54	          & 4.09	            & 339.10M     \\
          +CA(r=16)		    & MobileNetV2-1.0	   & 71.90           	 & \textbf{90.79} 	  & 4.37	            & 342.63M       \\
          +ECA		        & MobileNetV2-1.0	   & 71.04           	 & 90.19 	      	  & \textbf{3.51}	    & 329.49M       \\
          +SE		        & MobileNetV2-1.0	   & 71.11           	 & 90.14 	      	  & 3.79	            & 329.74M       \\
          +SA		        & MobileNetV2-1.0	   & 70.58           	 & 90.09 	          & \textbf{3.51}	    & \textbf{328.50M}       \\
          +CBAM		        & MobileNetV2-1.0	   & 71.24           	 & 90.42 	      	  & 3.79	            & 331.02M       \\
          +ELA-S(Ours)		& MobileNetV2-1.0	   & 71.52	             & 90.57              & 3.79	            & 334.97M      \\
          +ELA-L(Ours)		& MobileNetV2-1.0	   & \textbf{71.91}	     & 90.68              & 3.90	            & 337.94M       \\

         \hline
          ResNet18	        & ResNet18	          & 69.86	             & 89.08             & 11.690	         & 1.824G  \\
          +CA		        & ResNet18            & 69.77	              & 88.64	         & 11.886	         & 1.827G  \\
          +ECA		        & ResNet18            & 70.15	              & 89.45	         & \textbf{11.690}	 & 1.826G  \\
          +SE	            & ResNet18	          & 70.41	              & 89.65	         & 11.755	         & 1.826G  \\
          +CBAM	            & ResNet18	          & 70.48	              & 89.62	         & 11.756	         & 1.829G  \\
          +SA	            & ResNet18	          & 70.42	              & 89.52	         & 11.690	         & \textbf{1.825G}  \\
          +ELA-T(Ours)	    & ResNet18	          & 70.68	              & 89.68	         & 11.704	         & \textbf{1.825G}  \\
          +ELA-B(Ours)	    & ResNet18	          & \textbf{70.79}	      & \textbf{89.75}	 & 11.710	         & \textbf{1.825G}  \\

          \hline
          ResNet50	        & ResNet50	          & 75.83	              & 92.81	         & 25.557	         & 4.134G  \\
          +CA		        & ResNet50	          & 76.35	              & 93.06	         & 26.080	         & 4.141G  \\
          +ECA		        & ResNet50	          & 76.24	              & 93.17	         & 25.557	         & 4.136G  \\
          +SE		        & ResNet50	          & 76.01	              & 92.15	         & 25.905	         & 4.136G  \\
          +CBAM		        & ResNet50	          & 76.09	              & 92.90	         & 25.906	         & 4.138G  \\
          +SA		        & ResNet50	          & 76.30	              & 93.08	         & \textbf{25.557}	 & \textbf{4.135G}  \\
          +ELA-S(Ours)		& ResNet50	          & 76.59	              & 93.34	         & 25.713	         & 4.139G  \\
          +ELA-B(Ours)		& ResNet50            & \textbf{76.63}   	  & \textbf{93.35}	 & 25.565	         & \textbf{4.135G}  \\
          \hline
          ResNet101	        & ResNet101	          & 77.35	              & 93.56	         & 44.549	         & 7.866G  \\
          +CA		        & ResNet101	          & 77.42	              & 93.68	         & 45.595	         & 7.880G  \\
          +ELA-B(Ours)		& ResNet101	          & 77.41	              & 93.73	         & \textbf{44.705}	 & \textbf{7.871G}  \\
          +ELA-S(Ours)		& ResNet101           & \textbf{77.56}	      & \textbf{93.91}	 & 44.768	         & 7.874G  \\
          \hline
    \end{tabular}
    \label{tab5}
\end{table*}

\subsection*{4.2 Image Classification on ImageNet}
\textbf{MobileNetV2} We explored the performance of our proposed ELA-S, ELA-L, and CA using the MobilenetV2 architecture as the backbone. For the CA, we selected dimensionality reduction factors of 16, 24, and 32, respectively. In the validation experiments, we inserted the attention blocks into the MobileNetV2 Inverted Residual Block, ensuring consistent training settings for all models. The results presented in Table 4 demonstrate that our ELA-S module enhances the top-1 accuracy of the original MobileNetV2 by approximately 2.39\%. Furthermore, when compared to the CA method, our ELA method exhibits superior performance while utilizing fewer parameters and showcasing lower computational complexity. These findings further validate the efficiency and effectiveness of the ELA method.
\par
It is worth emphasizing that the inclusion of the CA (r=16) in MobileNetV2 significantly increases the parameter count by over 35\%. This poses a potential drawback for mobile networks as they tend to prioritize parameter efficiency.
\par
\textbf{ResNet} We compare our ELA  with various state-of-the-art attention modules, including SE block \cite{senet}, CBAM \cite{woo2018cbam}, ECA-Net \cite{wang2020eca}, SA-Net \cite{zhang2021sa}, and CA \cite{hou2021coordinate}. The evaluation metrics cover both efficiency (network parameters, floating point operations per second (GFLOPs)) and effectiveness (Top-1/Top-5 accuracy). To assess the efficiency and effectiveness of the different models, we employed the publicly available ResNet network with individual attentions incorporated in BasicBlock or BottleBlock, and applied them on the same computing platform. Compared to the state-of-the-art (SOTA) models, ELA achieves higher accuracy. Specifically, when ResNet18 serves as the backbone, ELA improves the Top-1 accuracy by 0.93\%. For ResNet50, the improvement is 0.8\%, and for ResNet-101, it is 0.21\%. Remarkably, when ResNet50 is used as the backbone, ELA only increases the number of parameters by 0.03\%, yet it enhances the absolute performance by 0.8\%, clearly demonstrating the effectiveness of ELA .
\par
We observed that the shortcomings of the CA become apparent when ResNet18 is utilized as the backbone. Its effectiveness is primarily demonstrated when employed in larger networks. On the other hand, ECA’s effectiveness is limited in this case, as the ResNet18 model has several layers with fewer than 128 channels. This mismatch with the ECA method’s settings \cite{wang2020eca}, which requires a 1D convolution kernel of 3, hinders its ability to effectively facilitate channel interaction.

\begin{table*}[h] 
    \centering
    \caption{Comparison of ELA and CA methods in COCO val2017. YOLOF is used as the baseline, with a dimensionality reduction factor of 32 for CA. The best records are represented in bold font.}
    \begin{tabular}{ c | c | c | c | c | c | c | c  }  
     \hline  
         Model	       & $AP_{50:95}(\%)$  & $AP_{50}(\%)$ 	& $AP_{75}(\%)$	  & $AP_S(\%)$	    & $AP_M(\%)$	  & $AP_L(\%)$	      & Params Size(MB) \\
         \hline
         YOLOF	        & 37.13	         & 56.7	            & 39.6	          & 17.8	        & 41.8	          & 51.4	          & 44.11   \\
         +CA	        & 37.45          & 57.3	            & 40.1	          & \textbf{18.8}	& 42.2	          & 51.8	          & 44.60   \\
         +ELA-L(Ours)  & \textbf{37.81}  &\textbf{57.9}     & \textbf{40.2}	  & \textbf{18.8}	& \textbf{42.6}	  & \textbf{52.4}	  & \textbf{44.34}   \\
         \hline
    \end{tabular}
    \label{tab6}
\end{table*}

\begin{table}[h]\small  
    \setlength{\tabcolsep}{2.5pt}
    \renewcommand{\arraystretch}{0.85}
    \caption{Results of testing different attention methods on the Pascal Voc2007 dataset using YOLOX-Nano as a baseline. It contains modules for channel downscaling, with the downscaling factor set to the usual 32. where Inf denotes Inference, the best records are indicated in bold.}
    \begin{tabular}{ c | c | c | c | c }  
     \hline  
         Method             & mAP               & Inf(FPS)              &\#P(MB)            & GFLOPs(G)    \\
         \hline
         YOLOX-Nano	        & 73.26	            & 82.87	               & 3.43	                   & 1.285    \\
         +CA	            & 72.62	            & 70.80	               & 3.52	                   & 1.286    \\
         +SE	            & 73.58	            & 77.28	               & 3.55	                   & 1.287    \\
         +CBAM	            & 73.80	            & 69.30	               & 3.56	                   & 1.289    \\
         +ECA	            & 73.38	            & \textbf{77.73}	   & \textbf{3.44}	          & 1.287    \\
         +SA	            & 73.35	            & 74.60	               & \textbf{3.44}	          & 1.286    \\
         +ELA-S(Ours)	    & \textbf{74.36}	& 75.68	               & 3.50	                   & \textbf{1.285}    \\
         \hline
    \end{tabular}
    \label{tab7}
\end{table}

\begin{table}[h]\small  
    \setlength{\tabcolsep}{2.5pt}
    \renewcommand{\arraystretch}{0.85}
    \caption{Semantic segmentation results of various attention methods using DeepLabV3 in Pascal VOC 2012. The output step size of the segmentation network in the experiment is 16, which refers to the output step size of the segmentation network. The module of channel dimensionality reduction is included, and the dimensionality reduction factor r is set to the usual 32. the best records are indicated in bold.}
    \begin{tabular}{ c | c | c | c  }  
     \hline  
         Method             & BackBones      & Mean IoU      & Params(MB)     \\
         \hline
         DeepLabV3	        & ResNet50	      & 70.47	        & 39.639     \\
         +SE			    & ResNet50        & 71.13	        & 39.796     \\
         +CA			    & ResNet50        & 71.17	        & 39.875     \\
         +CBAM			    & ResNet50        & 71.06	        & 39.719     \\
         +ECA			    & ResNet50        & 70.94	        & \textbf{39.642}     \\
         +SA			    & ResNet50        & 70.50	        & \textbf{39.642}     \\
         +ELA-S(Ours)		& ResNet50        & \textbf{71.55}  & 39.790     \\
         \hline
    \end{tabular}
    \label{tab8}
\end{table}

\subsection*{4.3 Object Detection}
\textbf{Object Detection on MS COCO}
\par
We have chosen the YOLOF \cite{2021yolof} object detector to compare the performance of the ELA method with the CA method. YOLOF utilizes ResNet50 \cite{he2016resnet} as the backbone and conducts experiments on the MS COCO \cite{2014coco} dataset. The Attention module is employed only five times, at the output of the backbone network, after the projecter of the Encoder, and after its first three Residual Blocks. For all experiments, the SGD optimizer was used. Training was done at a resolution of $800\times1333$ with a single scale, and each mini-batch contained a total of 16 images. The initial learning rate was set to $0.003$, with a warm-up period of 1500 iterations. A total of 12 epochs were trained according to the “1× schedule” scheme. The learning rate was reduced by a factor of 10 after the 8th and 11th epochs respectively. During model inference, the results were post-processed using an NMS with a threshold of 0.6. The remaining hyperparameters followed the settings specified in \cite{2021yolof}.
\par
According to Table \ref{tab6}, While the CA method is beneficial for enhancing the performance of the YOLOF object detector, our proposed ELA method surpasses the CA method significantly in terms of model parameters and the enhancement of diverse performance metrics. In particular, when employing YOLOF \cite{2021yolof} as the foundational detector (with ResNet-50 as the Backbone), ELA demonstrates a 0.68\% improvement in AP50:95.
\par
\textbf{Object Detection on Pascal VOC2007}
Furthermore, we validated the effectiveness of ELA using YOLOX-Nano \cite{ge2021yolox} on the Pascal VOC dataset. We added the ELA module and other corresponding attention modules only after the backbone of YOLOX-Nano and after the feature fusion layer in the neck. The model was trained using transfer learning by loading pre-trained weights from MS COCO. For all experiments, we utilized the SGD optimizer with an initial learning rate of 0.02. The learning rate decay was controlled using the cosine annealing strategy \cite{he2019bag}, with a weight decay of 0.0005 and a momentum of 0.937. The batch size was set to 16, and the input image size was $640\times640$. The model was trained for a total of 72 epochs on the training dataset. We excluded the mix-up augmentation and adjusted the mosaic technique by reducing the scale range from $[0.1, 2.0]$ to $[0.5, 1.5]$. For other hyperparameter configurations, please refer to \cite{ge2021yolox}.
\par
In Table \ref{tab7}, using the YOLOX-nano model as the baseline, we present the detection results on the Pascal VOC 2007 test set with different attention methods. We observed that the CA method significantly decreased the performance of the baseline, while ECA-Net and SA-Net had minimal improvement on the baseline performance. In contrast, incorporating our ELA led to a remarkable enhancement in the baseline performance, resulting in a 1.1\% mAP improvement. Both detection experiments on the MS COCO and Pascal VOC datasets indicated that the ELA method exhibited superior performance enhancement compared to other attention methods.

\subsection*{4.4 Semantic Segmentation}
\par
Finally, we present the semantic segmentation results obtained using the ELA of DeepLabV3 \cite{2017deeplabv3} on Pascal VOC2012 \cite{2015voc}. We utilize ResNet-50 \cite{he2016resnet} as the backbone, applying attention methods after each $3\times3$ convolution in every Bottleneck and after each layer. We employ the output $stride = 16$ approach and compute batch normalization statistics with a $batch\_size$ of 12. We use a cropping size of 513 and a decay of 0.9997 for training the batch normalization parameters. The training process involves 40K iterations on the train_aug dataset with an initial learning rate of 0.025. All models are implemented using the PyTorch toolkit  $\footnote{https://github.com/VainF/DeepLabV3Plus-Pytorch}$.

\par
As shown in Table \ref{tab8}, the CA method demonstrates superior performance enhancement capability for DeepLabV3. However, the SE block and ECA-Net methods do not show significant improvement in the performance of DeepLabV3. Surprisingly, the SA-Net actually leads to a decline in the performance of the DeepLabV3 model, indicating that the generalization ability of SA-Net is not particularly good. Compared to all the aforementioned methods, the ELA method exhibits the best performance improvement for the model. The experimental results presented above sufficiently demonstrate the strong generalization ability of the ELA method for various computer vision tasks.
\par

\section*{5. Conclusion}

This paper introduces an innovative attention mechanism called Efficient Local Attention (ELA), aiming to enhance the representation capability of CNNs. ELA simplifies the process of accurately localizing regions of interest with its lightweight and straightforward structure. Experimental results demonstrate that ELA is a plug-and-play attention method that does not necessitate channel dimensionality reduction. Moreover, ELA consistently achieves significant performance improvements across a range of deep CNN architectures.

\bibliographystyle{plain}
\bibliography{refer2}

\end{document}